%% file: main.tex
\documentclass[a4paper,11pt]{article}
\usepackage[margin=2cm]{geometry}
\usepackage[T1]{fontenc}
\usepackage[utf8]{inputenc}
\usepackage{lmodern}
\usepackage[british]{babel}
\usepackage{graphicx}
\usepackage{amsmath,amssymb,booktabs,tabularx,array}
\usepackage[numbers,sort&compress]{natbib}
\usepackage{siunitx}
\usepackage{pgfplots,pgfplotstable}
\pgfplotsset{compat=1.18}
\usepgfplotslibrary{groupplots}
\definecolor{paperblue}{HTML}{2a78d6}
\definecolor{paperorange}{HTML}{eb6834}
\pgfplotsset{
  paperaxis/.style={font=\sffamily\scriptsize,tick label style={font=\sffamily\scriptsize},
    title style={font=\sffamily\small},axis line style={black!40,thin},
    tick style={black!40,thin},ymajorgrids=true,grid style={black!10,thin},
    legend style={draw=none,fill=white},mark size=2pt},
  bluecurve/.style={color=paperblue,line width=0.7pt},
  orangecurve/.style={color=paperorange,line width=0.7pt}
}
\usepackage[unicode,hidelinks]{hyperref}
\usepackage{url}
\newcommand{\RadiusImages}{50}

\pgfplotstableread[col sep=comma]{generated/recovery-fcn.csv}\RecoveryFCN
\pgfplotstableread[col sep=comma]{generated/recovery-deeplab.csv}\RecoveryDeepLab
\newcommand{\DataValue}[3]{%
  \pgfplotstablegetelem{#2}{#3}\of{#1}%
  \expandafter\num\expandafter{\pgfplotsretval}}

\hypersetup{pdftitle={Limits of local curvature for constructing search subspaces in segmentation models},pdfauthor={Maksim Antonovich Bolshim and Alexander Vladimirovich Kugaevskikh}}
\begin{document}
\title{Limits of local curvature for constructing search subspaces in segmentation models}
\author{M.A.~Bolshim*, A.V.~Kugaevskikh}
\date{}
\maketitle
\begin{center}\small ITMO University, 49 Kronverkskiy Prospekt, Bldg.~A, Saint Petersburg, 197101, Russia\\[0.5em]mabolshim@itmo.ru (M.A. Bolshim), avkugaevskikh@itmo.ru (A.V. Kugaevskikh)\\*Corresponding author\end{center}
\begin{abstract}
We examine how well subspaces constructed from local curvature represent the change in the gradient update field after a finite step in semantic segmentation models. The local response is the action of the curvature operator on the actual input displacement. In a study of 50 images and two models, we measure the ratio of the loss increase attained in a fixed subspace to that attained in the full space. This ratio decreases as the search radius increases. On ten images from the same sample, we compare the one-dimensional subspace spanned by the initial gradient with two rank-16 subspaces for both models. The two equal-rank subspaces are assessed at the same initial states, using the same actual steps and measured vectors. Adding generalised Gauss--Newton directions to the initial gradient reduces the mean relative error in representing the field change from 0.922 to 0.874 for the first model and from 0.819 to 0.763 for the second. At the same rank, including the response direction for the initial gradient nearly eliminates the error in representing the local response. The mean error in representing the field change does not decrease. A decomposition with fixed pixel weights shows that the finite-step remainder is comparable in norm to the field change and makes the largest diagonal contribution to the squared projection error of the field after the step. Evaluation of a search subspace should measure how accurately it represents the field change and the local response.
\end{abstract}

\normalsize

\input{article}
\renewcommand{\refname}{References}
\input{references}
\input{supplement}
\end{document}

%% file: article.tex
\section*{Introduction}

Subspaces constructed from the gradient and local curvature can reduce the dimensionality of the search for input perturbations in semantic segmentation models. As the input changes, both the magnitude and direction of the gradient change. The action of the curvature operator on the input displacement defines a local response. We ask whether a subspace that accurately represents this response also represents the actual change in the gradient update field after a finite step.

The vulnerability of segmentation models to deliberately constructed perturbations has been studied in \citep{xie2017adversarial,arnab2018robustness}. In pixel-wise prediction, directions of loss change for different image regions may compete. Methods with adaptive weights, such as CosPGD \citep{agnihotri2024cospgd}, account for this heterogeneity. The gradient update field studied here depends both on the loss function and on the rule for assigning pixel weights.

Krylov subspaces allow neural network curvature to be used without constructing a dense matrix \citep{vinyals2012krylov}. The generalised Gauss--Newton matrix (GGN) characterises the local loss model \citep{martens2020natural}. Related work has examined the relationship between input-space loss curvature and robustness \citep{moosavi2019curvature}, and the alignment of low-dimensional perturbation spaces with gradients \citep{dohmatob2023origins}.

We use two experiments on two semantic segmentation models to examine this question. The radius study measures the loss increase attained in a fixed subspace as the search region expands. A paired diagnostic compares two equal-rank subspaces at the same initial states, using the same actual steps and measured vectors. One of these subspaces includes the GGN response direction for the initial gradient. Although this nearly eliminates the error in representing the local response, it does not reduce the mean error for the field change. We separate the field change into the local response and a finite-step remainder to identify their contributions to the projection error.

\section{Materials and methods}\label{sec:methods}

\subsection{Models, data and coordinates}\label{sec:models}

Two models from the Torchvision library are used: a fully convolutional network (FCN) based on ResNet50 \citep{long2015fully} and DeepLabV3--MobileNetV3-Large. The DeepLabV3 feature extraction scheme is described in Section 3.1 of \citep{chen2018encoder}, and the MobileNetV3 architecture in \citep{howard2019mobilenetv3}. Both use the \nolinkurl{COCO_WITH_VOC_LABELS_V1} pretrained weights for the full segmentation model, and run in evaluation mode without activation smoothing.

Images and annotations are taken from the validation split of the PASCAL Visual Object Classes (VOC) 2012 dataset \citep{everingham2010pascal}. Images are resized to 320 \(\times\) 320 pixels using bilinear interpolation, and annotations using nearest-neighbour interpolation. The set $V$ contains pixels with valid annotations; label 255 is excluded, and $N=|V|$ is fixed for each image.

The input coordinates are defined as ${x}_{c}=({I}_{c}-{\mu }_{c})/{\sigma }_{c}$, where ${I}_{c}\in [0,\allowbreak 1]$ is the original intensity, $\mu =(0.485;\allowbreak \,0.456;\allowbreak \,0.406)$ and $\sigma =(0.229;\allowbreak \,0.224;\allowbreak \,0.225)$. Admissible perturbations satisfy $\|x-{x}_{\mathrm{clean}}{\|}_{2}\leq \varepsilon $, without subsequent clipping of red, green and blue (RGB) channel values. Thus, the geometry of the continuous normalised input is studied.

The radius study uses 50 images with indices $120,\allowbreak \ldots ,\allowbreak 169$ in the official VOC validation list, using zero-based indexing. Every fifth image was selected in advance for the diagnostic: indices $120,\allowbreak 125,\allowbreak \ldots ,\allowbreak 165$. These ten images are part of the original sample, with one observation per model for each image, giving 20 image--model pairs. Table~\ref{tab:protocol} summarises the experiments; \hyperref[app:details]{Appendix A} gives image identifiers and numerical details.

\begin{table}[!htbp]
\centering\small
\caption{Parameters of the two experiments. Both use two segmentation models and images of 320 \(\times\) 320 pixels}
\label{tab:protocol}
\begin{tabularx}{\linewidth}{@{}p{0.14\linewidth}XX@{}}
\toprule
Parameter & Radius study & Post-step field diagnostic \\
\midrule
Images & 50 VOC 2012 images, indices $120,\allowbreak \ldots ,\allowbreak 169$ & Every fifth image in the same sample: 10 images, two models \\
Loss & Mean cross-entropy & Cross-entropy with cosine weights held fixed during differentiation \\
Radius & $\varepsilon \approx \{0.0625;\allowbreak \,0.25;\allowbreak \,1;\allowbreak \,4;\allowbreak \,64\}$ & $R\approx 4$; first-step length approximately 0.0391 \\
Subspace & Gradient and GGN Ritz directions; ranks from 1 to 64 & ${A}_{\mathrm{ref}}$ and $B$: rank 16; gradient control: rank 1 \\
Construction & Up to 80 Krylov directions; basis fixed at the initial point & Up to 80 Krylov directions; two bases at the same initial point \\
Comparison & 64 steps from a zero start or 256 steps from three starts & State, actual step and measured vectors shared by both subspaces \\
Metric & Recovery of the loss increase relative to the attained full-space reference & Relative representation error for the response, field after the step and field change \\
\bottomrule
\end{tabularx}
\end{table}

\subsection{Local curvature and the update field}\label{sec:curvature}

Let ${z}_{i}(x)$ denote the logits at pixel $i$, ${p}_{i}(x)=\operatorname{softmax}\,{z}_{i}(x)$, ${y}_{i}$ the true class, and ${J}_{i}=\partial {z}_{i}/\partial x$. Denote cross-entropy by $\operatorname{CE}$. For fixed non-negative weights ${a}_{i}$, the loss function and its GGN are

\begin{equation}
{L}_{a}(x)=\frac{1}{N}\sum_{i\in V}{a}_{i}\,\operatorname{CE}({z}_{i}(x),\allowbreak {y}_{i}),\allowbreak 
\label{eq:loss}
\end{equation}

\begin{equation}
{G}_{a}(x)=\frac{1}{N}\sum_{i\in V}{J}_{i}^{T}{a}_{i}(\operatorname{diag}\,{p}_{i}-{p}_{i}{p}_{i}^{T}){J}_{i}.
\label{eq:ggn}
\end{equation}

Here, local curvature means the GGN operator \eqref{eq:ggn}, not the full input Hessian. It combines the curvature of cross-entropy with respect to the logits with the first derivatives of the logits with respect to the input. Spectral directions are approximated by the Rayleigh--Ritz method in a Krylov subspace, without forming a dense matrix \citep{parlett1998symmetric}.

In the radius study, ${a}_{i}=1$. In the diagnostic, the weight ${a}_{i}(x)$ is the cosine similarity between ${p}_{i}(x)$ and the unit vector of the true class. The norm in the denominator has a lower bound of ${10}^{-8}$. This weighting rule is based on CosPGD \citep{agnihotri2024cospgd}. The weights are held fixed when computing the gradient in this diagnostic. For the transition ${x}_{0}\to {x}_{1}$, let ${a}^{0}=a({x}_{0})$, ${a}^{1}=a({x}_{1})$, and define three fields:

\begin{equation}
{f}_{0}=\nabla {L}_{{a}^{0}}({x}_{0}),\allowbreak \quad \quad {f}_{1}^{\mathrm{fix}}=\nabla {L}_{{a}^{0}}({x}_{1}),\allowbreak \quad \quad {f}_{1}^{\mathrm{dyn}}=\nabla {L}_{{a}^{1}}({x}_{1}).
\label{eq:fields}
\end{equation}

The superscripts $\mathrm{fix}$ and $\mathrm{dyn}$ distinguish weights retained from ${x}_{0}$ from those recomputed at ${x}_{1}$; in both cases the weights are held fixed during differentiation.

\subsection{Radius study}\label{sec:radius}

A subspace comprising the gradient and GGN Ritz directions is constructed at the initial point and held fixed. Up to 80 Krylov directions are constructed to approximate the operator's eigendirections. Ranks $k\in \{1,\allowbreak 2,\allowbreak 4,\allowbreak 8,\allowbreak 16,\allowbreak 32,\allowbreak 64\}$ are tested; linearly dependent directions are discarded without random completion. We test five radii, $\varepsilon \approx \{0.0625;\allowbreak \,0.25;\allowbreak \,1;\allowbreak \,4;\allowbreak \,64\}$, with exact values given in \hyperref[app:images]{Appendix A.1}.

The search uses projected gradient ascent with the direction normalised by its Euclidean norm. The baseline solver uses 64 steps starting from zero perturbation; the enhanced solver uses 256 steps, a zero start and two random starts in the coefficients of the same basis. The step length is $2.5\varepsilon /T$, where $T$ is the number of steps. The multiplier 2.5 was selected from $\{1.25;\allowbreak \,2.5;\allowbreak \,5\}$ on the first five images of the VOC training split and was not tuned on the validation sample.

For each image, model and radius, the full-space reference is the highest CE attained among runs with $T\in \{64,\allowbreak 128,\allowbreak 256\}$, one zero start and two random starts; the best point along the trajectory is retained. The loss-increase recovery ratio in subspace ${S}_{k}$ is defined as

\begin{equation}
\mathcal R({S}_{k})=\frac{L({x}_{{S}_{k}}^{\mathrm{best}})-L({x}_{\mathrm{clean}})}{L({x}_{\mathrm{full}}^{\mathrm{best}})-L({x}_{\mathrm{clean}})}.
\label{eq:recovery}
\end{equation}

Here, $L$ is the mean cross-entropy. Values above one are retained, since the full-space reference is an attained value rather than a global optimum. We report the median and interquartile range at $k=64$, for which all observed bases have the full rank of 64, and separately count images reaching the 0.9 threshold at any tested rank.

\subsection{Paired diagnostic of the field after the step}\label{sec:paired}

The diagnostic uses $R\approx 4$ and step length $\alpha =2.5R/256\approx 0.0391$. Two rank-16 subspaces are constructed at the initial point. Subspace ${A}_{\mathrm{ref}}$ contains the gradient and 15 Ritz directions; $B$ contains the gradient, an independent $Gq$ direction and 14 Ritz directions. Here, $G={G}_{{a}^{0}}({x}_{0})$, and $q$ is the normalised gradient of the weighted sum of losses at ${x}_{0}$. Ritz directions are ordered by decreasing approximate eigenvalue, with linearly dependent directions discarded. The subspaces have the same rank but are not nested.

The first step is taken in subspace $B$. If ${Q}_{B}$ contains its basis vectors as rows, the coefficients $c={Q}_{B}{g}_{0}$ and the direction ${Q}_{B}^{T}c/\operatorname{max}(\|c{\|}_{2},\allowbreak {10}^{-12})$ are computed, where ${g}_{0}$ is the gradient of the sum of losses. After taking a step of length $\alpha$, we check that the updated input lies in the admissible ball. We apply both projectors to the same vectors measured for the transition ${x}_{0}\to {x}_{1}$, using the actual displacement $s={x}_{1}-{x}_{0}$ in all measurements.

The local response to this displacement is defined by the action of the GGN operator:

\begin{equation}
h={G}_{{a}^{0}}({x}_{0})s.
\label{eq:response}
\end{equation}

For a basis with orthonormal rows, the projection is $Pv={Q}_{P}^{T}{Q}_{P}v$. The relative representation error for a non-zero vector is

\begin{equation}
{\rho }_{P}(v)=\frac{\|(I-P)v{\|}_{2}}{\|v{\|}_{2}}.
\label{eq:rho}
\end{equation}

The fraction of the squared vector norm captured by the subspace is $1-\rho_P(v)^2$. Errors are measured for $h$, ${f}_{1}^{\mathrm{fix}}$ and ${f}_{1}^{\mathrm{dyn}}$. The error difference $\Delta (v)={\rho }_{{A}_{\mathrm{ref}}}(v)-{\rho }_{B}(v)$ is positive when $B$ has the smaller error.

As a control, we use the one-dimensional subspace of the initial gradient: ${P}_{g}v={e}_{0}\langle {e}_{0},\allowbreak v\rangle $, ${e}_{0}={f}_{0}/\|{f}_{0}\|$. Comparing it with the curvature subspaces measures the combined effect of adding directions and increasing the rank. To isolate the change after the step, we also evaluate ${d}^{\mathrm{fix}}={f}_{1}^{\mathrm{fix}}-{f}_{0}$ and, analogously, ${d}^{\mathrm{dyn}}$. We compare the scale of the response projection residual with both the full field and its change by dividing its norm by $\|{f}_{1}^{\mathrm{fix}}\|$ and $\|{d}^{\mathrm{fix}}\|$, respectively.

The analysis concerns the first step in each of the 20 image--model pairs. We estimate intervals for the mean error difference $\Delta$ using a paired bootstrap with images as clusters: each of 2000 resamples draws ten images with replacement, retaining observations for both models and both projectors whenever an image is selected. The same resamples are used for fixed and recomputed weights. The 95\% interval bounds are the 2.5th and 97.5th percentiles, calculated with linear interpolation.

\subsection{The finite-step remainder and its projection}\label{sec:remainder}

The difference between the fixed-weight field change and the local response defines the finite-step remainder $u$. Recomputing the weights contributes a further term $w$:

\begin{equation}
u={f}_{1}^{\mathrm{fix}}-{f}_{0}-h,\allowbreak \quad \quad w={f}_{1}^{\mathrm{dyn}}-{f}_{1}^{\mathrm{fix}}.
\label{eq:remainders}
\end{equation}

Then ${f}_{1}^{\mathrm{fix}}={f}_{0}+h+u$, while the field with recomputed weights additionally contains $w$. The remainder $u$ accounts for the difference between the GGN and the derivative of the field, the nonlinear change in the field over a finite step, and numerical errors. The scale of the remainder is measured by the ratio $\|u\|/\|{d}^{\mathrm{fix}}\|$.

For $c=({f}_{0},\allowbreak h,\allowbreak u)$ and ${r}_{i}=(I-P){c}_{i}$,

\begin{equation}
\|(I-P){f}_{1}^{\mathrm{fix}}{\|}^{2}=\sum_{i}\|{r}_{i}{\|}^{2}+\sum_{i<j}2\langle {r}_{i},\allowbreak {r}_{j}\rangle .
\label{eq:balance}
\end{equation}

For each image--model pair, all terms are divided by $\|{f}_{1}^{\mathrm{fix}}{\|}^{2}$ before averaging. The mean diagonal and signed cross terms therefore sum to the mean squared relative error, ${\rho }_{P}{({f}_{1}^{\mathrm{fix}})}^{2}$.

\section{Results}\label{sec:results}

\subsection{Recovery ratio at different search radii}

\begin{figure}[tb]
\centering
\includegraphics[width=\linewidth]{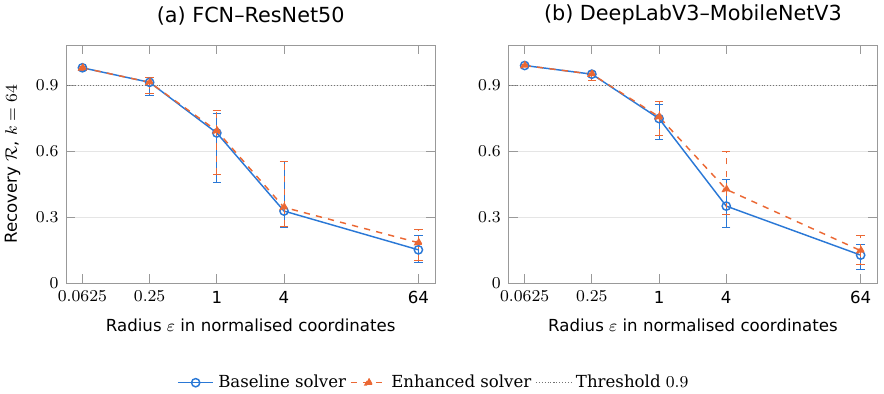}
\caption{(a) FCN--ResNet50; (b) DeepLabV3--MobileNetV3. Recovery of the cross-entropy increase in a fixed rank-64 subspace constructed from the gradient and curvature: the median and interquartile range over \RadiusImages{} images. The reference is the best full-space result attained for the same image, model and radius. The range characterises the spread of values across images. The curves almost coincide at small radii.}
\label{fig:recovery}
\end{figure}

Figure~\ref{fig:recovery} shows loss-increase recovery at $k=64$. At the smallest radius, the median is close to one: 0.979 for FCN and 0.989 for DeepLab with the baseline solver. For both solvers, the 0.9 threshold is attained on 48 of 50 FCN images and all 50 DeepLab images. Considering all tested ranks gives the same number of images reaching the threshold as rank 64 alone.

At the largest radius, the medians decrease to 0.151 for FCN and 0.127 for DeepLab with the baseline solver. Increasing the number of iterations and initialisations in the enhanced solver raises these values to 0.183 and 0.147, respectively. The 0.9 threshold is not attained for any image at any tested rank.

\subsection{Representation errors for the response and the field after the step}

The mean representation error for the response $h$ decreases from $4.26\times {10}^{-3}$ in ${A}_{\mathrm{ref}}$ to $7.44\times {10}^{-7}$ in $B$, with a smaller error in all 20 cases.

\begin{figure}[tb]
\centering
\includegraphics[width=\linewidth]{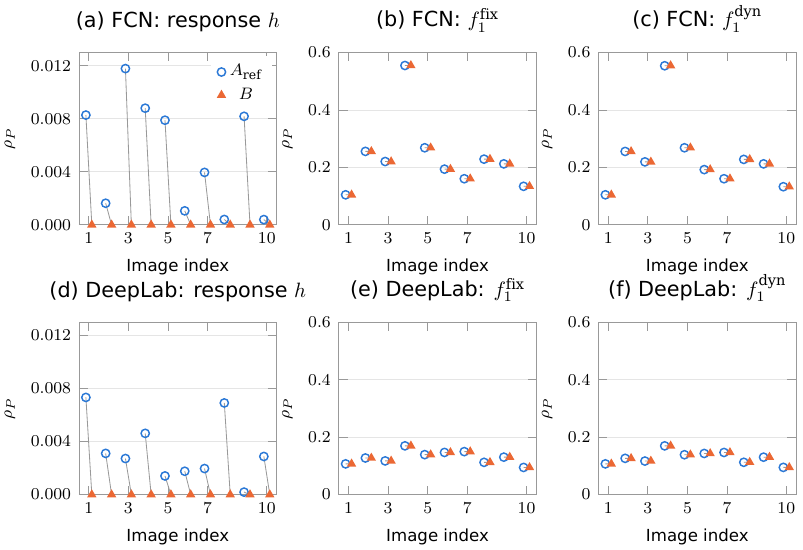}
\caption{Representation errors for the local response $h$ and the field after the step with fixed ($f_1^{\rm fix}$) and recomputed ($f_1^{\rm dyn}$) weights. The upper row (a--c) corresponds to FCN and the lower row (d--f) to DeepLab. Segments connect the values for $A_{\rm ref}$ (circle on the left) and $B$ (triangle on the right) for the same image. Image numbers are consistent across all panels. A common scale is used for the field after the step.}
\label{fig:response}
\end{figure}

For the field after the step, the mean errors in ${A}_{\mathrm{ref}}$ and $B$ are 0.18136 and 0.18139 with fixed weights, and 0.18063 and 0.18072 with recomputed weights. Subspace $B$ has the smaller error in 5 and 4 of 20 cases, respectively. With fixed weights, the mean error difference is $-3.69\times {10}^{-5}$, and the 95\% interval is $[-8.08\times {10}^{-5};\allowbreak \;2.28\times {10}^{-5}]$, which includes zero. With recomputed weights, the mean error difference is $-9.51\times {10}^{-5}$, and the interval is $[-1.51\times {10}^{-4};\allowbreak \;-4.84\times {10}^{-5}]$. In both cases, the absolute mean error difference is less than ${10}^{-4}$. Figure~\ref{fig:response} shows the errors for all 20 pairs.

\subsection{Comparison with the one-dimensional gradient subspace}

Table~\ref{tab:controls} compares errors for the three subspaces. Moving from the one-dimensional gradient subspace to ${A}_{\mathrm{ref}}$ reduces the mean representation error for $h$ from 0.359 to 0.00524 for FCN and from 0.373 to 0.00329 for DeepLab. The error in representing the field change also decreases, from 0.922 to 0.874 and from 0.819 to 0.763, respectively. The error reduction for the field change is smaller than that for the local response.

\begin{table}[!htbp]
\centering\small
\caption{Gradient control and two curvature subspaces. Values are means of individual ratios over ten images per model; ${d}^{\mathrm{fix}}={f}_{1}^{\mathrm{fix}}-{f}_{0}$, ${e}_{h}=\|(I-P)h\|$. The gradient control has rank 1, and subspaces ${A}_{\mathrm{ref}}$ and $B$ have rank 16}
\label{tab:controls}
\begin{tabular}{@{}llllll@{}}
\toprule
Model & Subspace & $\rho (h)$ & $\rho ({f}_{1}^{\mathrm{fix}})$ & $\rho ({d}^{\mathrm{fix}})$ & ${e}_{h}/\|{d}^{\mathrm{fix}}\|$ \\
\midrule
FCN & Gradient & 0.3586 & 0.2457 & 0.9224 & 0.1878 \\
FCN & ${A}_{\mathrm{ref}}$ & $5.236\times {10}^{-3}$ & 0.2332 & 0.8741 & $2.365\times {10}^{-3}$ \\
FCN & $B$ & $8.239\times {10}^{-7}$ & 0.2333 & 0.8742 & $4.165\times {10}^{-7}$ \\
DeepLab & Gradient & 0.3734 & 0.1393 & 0.8185 & 0.2144 \\
DeepLab & ${A}_{\mathrm{ref}}$ & $3.290\times {10}^{-3}$ & 0.1295 & 0.7627 & $2.035\times {10}^{-3}$ \\
DeepLab & $B$ & $6.633\times {10}^{-7}$ & 0.1295 & 0.7630 & $3.933\times {10}^{-7}$ \\
\bottomrule
\end{tabular}
\end{table}

In ${A}_{\mathrm{ref}}$, the response $h$ is already represented with a small error. Replacing this subspace with $B$ reduces the norm of the response projection residual by a mean of 0.236\% of the norm of ${d}^{\mathrm{fix}}$ for FCN and 0.203\% for DeepLab; the maximum over all cases is 0.585\%. For the response itself, the mean ratios $\|h\|/\|{d}^{\mathrm{fix}}\|$ are 0.522 and 0.596.

With fixed weights, the mean errors in representing the field change in subspaces ${A}_{\mathrm{ref}}$ and $B$ are 0.8741 and 0.8742 for FCN, and 0.7627 and 0.7630 for DeepLab, respectively. With recomputed weights, the corresponding values are 0.8870 and 0.8874 for FCN, and 0.8212 and 0.8217 for DeepLab.

\subsection{Contribution of the finite-step remainder to the projection error}

The median ratio $\|u\|/\|{d}^{\mathrm{fix}}\|$ is 0.925, with a range from 0.567 to 1.16. Values above one arise from partial cancellation between the local response and the finite-step remainder. In norm, the finite-step remainder is comparable to the field change.

The component of the remainder $u$ orthogonal to the subspace gives the largest diagonal term in the squared error for the field after the step. In ${A}_{\mathrm{ref}}$, the cross term between the projection residuals of $h$ and $u$ is negative and reduces their combined energy. In subspace $B$, the contribution of $h$ decreases on average, but the diagonal contribution of $u$ increases and cancellation weakens.

\begin{figure}[tb]
\centering
\includegraphics[width=\linewidth]{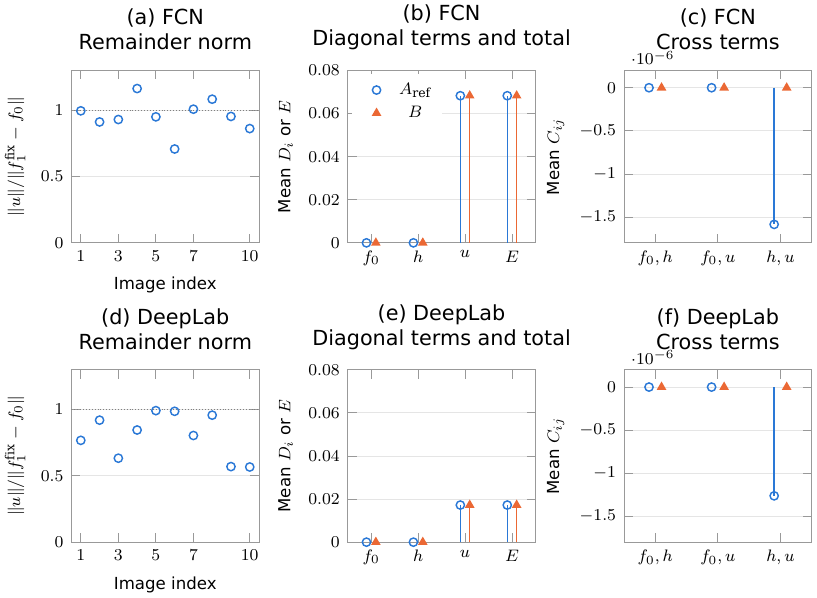}
\caption{Size of the finite-step remainder and decomposition of the squared error with fixed weights. The upper row (a--c) corresponds to FCN and the lower row (d--f) to DeepLab. Left: the ratio of norms for each image; the dotted line denotes unity. Centre: the mean normalised diagonal terms $D_i=\|(I-P)c_i\|^2/\|f_1^{\rm fix}\|^2$ and total $E=\rho_P(f_1^{\rm fix})^2$. Right: the signed terms $C_{ij}=2\langle(I-P)c_i,(I-P)c_j\rangle/\|f_1^{\rm fix}\|^2$, where $c=(f_0,h,u)$. Normalisation precedes averaging, $\overline E=\sum\overline D_i+\sum\overline C_{ij}$. The scales for the diagonal and cross terms differ.}
\label{fig:remainder}
\end{figure}

Moving from ${A}_{\mathrm{ref}}$ to $B$ increases the mean normalised error energy by $1.60\times {10}^{-5}$ for FCN and $1.17\times {10}^{-5}$ for DeepLab. The corresponding mean error energies in ${A}_{\mathrm{ref}}$ are 0.0683 and 0.0172. Figure~\ref{fig:remainder} shows the terms for both subspaces, and Table~\ref{tab:appendix-balance} gives their mean differences.

\section{Discussion}\label{sec:discussion}

All the subspaces studied contain the initial gradient, which explains why the full field is better represented than its change. In Table~\ref{tab:controls}, the error for the field after the step is approximately 0.13--0.23 in the curvature subspaces, whereas the field-change error reaches 0.76--0.87. Subtracting the initial field thus exposes a relative error that is less apparent when the full field is evaluated.

Including the $Gq$ direction reduces the norm of the local-response projection residual by an average of approximately 0.2\% of the norm of the field change. A substantial part of the change remains outside both subspaces and is associated with the finite-step remainder $u$. The field decomposition in \eqref{eq:remainders}--\eqref{eq:balance} identifies the contributions of the local response and the finite-step remainder to the projection error. Therefore, the representation errors for the local response and the field change should be considered together when evaluating a subspace.

The two experiments address different aspects of a fixed search subspace. The paired diagnostic examines the first step on ten images for the two fixed models; the radius study measures loss recovery as the search region expands, using unweighted cross-entropy and subspaces of several ranks. Its full-space reference is the best result attained with the specified numbers of steps and initialisations. Both experiments concern continuous normalised input coordinates.

\section*{Conclusion}

In the segmentation models studied, subspaces constructed from the gradient and curvature have smaller errors in representing the local response and the field change after the first step than the one-dimensional subspace of the initial gradient. When equal-rank subspaces are compared, nearly eliminating the error in representing the local response leaves the field-change error high. The finite-step remainder makes the largest diagonal contribution to the squared projection error of the field with fixed weights. The radius study shows a decrease in the loss-increase recovery ratio as the search region expands. For the evaluation of a local search subspace, these results support measuring the representation errors for both the local response and the actual field change.

%% file: references.tex
\begingroup
\setlength{\bibsep}{2pt}

\endgroup

%% file: supplement.tex
\appendix
\renewcommand{\thesection}{Appendix~\Alph{section}}
\renewcommand{\thesubsection}{\Alph{section}.\arabic{subsection}}
\renewcommand{\theHsection}{appendix.\arabic{section}}
\renewcommand{\theHsubsection}{\theHsection.\arabic{subsection}}
\setcounter{table}{0}
\renewcommand{\thetable}{\Alph{section}\arabic{table}}
\renewcommand{\theHtable}{\theHsection.\arabic{table}}
\section{Sample and numerical details}\label{app:details}

\subsection{Images and search parameters}\label{app:images}

Table~\ref{tab:appendix-images} identifies the ten diagnostic images in the order used in Figures~\ref{fig:response} and \ref{fig:remainder}. Indices are zero-based positions in the official VOC 2012 validation list, \nolinkurl{ImageSets/Segmentation/val.txt}.

\begin{table}[!htbp]
\centering\small
\caption{Images in the diagnostic sample}
\label{tab:appendix-images}
\begin{tabular}{@{}lll@{}}
\toprule
Number in figures & Index in VOC list & Image identifier \\
\midrule
1 & 120 & \nolinkurl{2007_003611} \\
2 & 125 & \nolinkurl{2007_003742} \\
3 & 130 & \nolinkurl{2007_003872} \\
4 & 135 & \nolinkurl{2007_004052} \\
5 & 140 & \nolinkurl{2007_004190} \\
6 & 145 & \nolinkurl{2007_004380} \\
7 & 150 & \nolinkurl{2007_004510} \\
8 & 155 & \nolinkurl{2007_004712} \\
9 & 160 & \nolinkurl{2007_004969} \\
10 & 165 & \nolinkurl{2007_005149} \\
\bottomrule
\end{tabular}
\end{table}

In the radius study, the exact radii are $\varepsilon =b/0.229$, where $b\in \{0.014313;\allowbreak \,0.057251;\allowbreak \,0.229004;\allowbreak \,0.916016;\allowbreak \,14.65625\}$. The diagnostic uses $R=4.0000698689956335$ and nominal step length $\alpha=0.039063182314410486$, starting from the unperturbed image. The random number generator seed is zero in both experiments. Since the accumulated displacement is zero at the initial point, no separate accumulated-displacement direction is included in the diagnostic basis.

\subsection{Basis construction and numerical arithmetic}\label{app:arithmetic}

In the diagnostic, the Krylov process starts from the normalised gradient obtained by differentiating the mean loss with fixed weights in 32-bit floating-point arithmetic (FP32). At each of up to 80 steps, the direction is orthogonalised against the retained directions. The process stops when the residual norm does not exceed ${10}^{-8}$. The eigenvalue problem for the symmetrised projection of the operator is solved in 64-bit floating-point arithmetic (FP64). Ritz directions are ordered by decreasing eigenvalue and converted to FP32.

The initial gradient is retained first in the final basis. In $B$, an independent $Gq$ direction is then included, followed by Ritz directions up to rank 16. In ${A}_{\mathrm{ref}}$, Ritz directions follow the gradient. The independence check and two orthogonalisation passes are performed in FP64. The independence threshold is ${10}^{-7}$ times the largest singular value of the matrix of candidate directions, whose rows are the normalised initial gradient, normalised $Gq$ for $B$, and the Ritz directions added sequentially. Projections use these bases converted to FP64, without rebuilding them.

The step coefficients and updated input are computed in FP32. For field analysis, the gradient $g$ of the sum of weighted losses is computed in FP32, then converted to FP64 and divided by the fixed number $N$ of valid pixels, not by the sum of weights. Thus the field and actual displacement used in the analysis are
\[
f=\operatorname{FP64}(g)/N,\qquad s=\operatorname{FP64}({x}_{1})-\operatorname{FP64}({x}_{0}).
\]
To evaluate the local response in \eqref{eq:response}, the displacement is split into two FP32 parts, $s={s}_{\mathrm{hi}}+{s}_{\mathrm{lo}}$. The two operator actions are converted to FP64 and summed:
\[
h=\operatorname{FP64}({G}_{{a}^{0}}^{\operatorname{FP32}}{s}_{\mathrm{hi}})+\operatorname{FP64}({G}_{{a}^{0}}^{\operatorname{FP32}}{s}_{\mathrm{lo}}).
\]
The GGN operator already includes the factor $1/N$; no further normalisation is applied to $h$. All projections and relative errors, including those for the gradient control, are evaluated for each image--model pair before averaging.

The diagnostic used PyTorch 2.12.0+\allowbreak{}cu130 and Torchvision 0.27.0 with CUDA 13.0 and cuDNN 92000. Matrix operations and cuDNN used IEEE FP32 arithmetic. Automatic cuDNN algorithm selection was disabled, and deterministic mode was enabled with a warning when no deterministic implementation was available. Bootstrap resamples were generated in Python 3.12.13 by successive calls to \nolinkurl{random.Random.choice}, with seed 20261029 and the ten image identifiers sorted in ascending order.

\subsection{Projection-error balance}\label{app:balance}

Table~\ref{tab:appendix-balance} gives the mean differences $B-{A}_{\mathrm{ref}}$ between the terms of \eqref{eq:balance}. Here ${r}_{v}=(I-P)v$ and ${E}_{P}={\rho }_{P}{({f}_{1}^{\mathrm{fix}})}^{2}$ are evaluated separately for each subspace. Each term is divided by $\|{f}_{1}^{\mathrm{fix}}\|^{2}$ for the same image--model pair before averaging. The last two rows give the mean error energies themselves rather than their differences.

\begin{table}[!htbp]
\centering\small
\caption{Mean normalised differences $B-{A}_{\mathrm{ref}}$ between projection-error terms, with mean error energies in the last two rows}
\label{tab:appendix-balance}
\begin{tabular}{@{}lll@{}}
\toprule
Term & FCN & DeepLab \\
\midrule
$\|{r}_{{f}_{0}}{\|}^{2}$ & $-1.13\times {10}^{-19}$ & $-3.03\times {10}^{-20}$ \\
$\|{r}_{h}{\|}^{2}$ & $-8.74\times {10}^{-7}$ & $-1.60\times {10}^{-7}$ \\
$\|{r}_{u}{\|}^{2}$ & $+1.53\times {10}^{-5}$ & $+1.06\times {10}^{-5}$ \\
$2\langle {r}_{{f}_{0}},\allowbreak {r}_{h}\rangle $ & $+5.57\times {10}^{-14}$ & $-3.97\times {10}^{-14}$ \\
$2\langle {r}_{{f}_{0}},\allowbreak {r}_{u}\rangle $ & $+8.03\times {10}^{-15}$ & $+4.07\times {10}^{-13}$ \\
$2\langle {r}_{h},\allowbreak {r}_{u}\rangle $ & $+1.59\times {10}^{-6}$ & $+1.26\times {10}^{-6}$ \\
${E}_{B}-{E}_{{A}_{\mathrm{ref}}}$ & $+1.59664\times {10}^{-5}$ & $+1.17307\times {10}^{-5}$ \\
${E}_{{A}_{\mathrm{ref}}}$ & 0.0682952 & 0.0172226 \\
${E}_{B}$ & 0.0683112 & 0.0172344 \\
\bottomrule
\end{tabular}
\end{table}

The reduction in the diagonal contribution of $h$ is accompanied by a larger contribution from the finite-step remainder and weaker cross-term cancellation. The resulting increases in mean normalised error energy, $1.59664\times {10}^{-5}$ for FCN and $1.17307\times {10}^{-5}$ for DeepLab, are small relative to the mean error energies in the last two rows.